\documentclass[conference]{IEEEtran}
\IEEEoverridecommandlockouts

\usepackage{cite}
\usepackage{amsmath,amssymb,amsfonts}
\usepackage{algorithmic}
\usepackage{graphicx}
\usepackage{textcomp}
\usepackage{xcolor}
\usepackage{enumerate}
 \usepackage{hyperref}
\usepackage{mathrsfs}
\usepackage{bm}
\usepackage{url}
\usepackage{xspace}
\usepackage{subcaption}
\usepackage{tabularray}

\def\BibTeX{{\rm B\kern-.05em{\sc i\kern-.025em b}\kern-.08em
    T\kern-.1667em\lower.7ex\hbox{E}\kern-.125emX}}
\begin{document}


\title{UOPSL: Unpaired OCT Predilection Sites Learning \\for Fundus Image Diagnosis Augmentation \\
}
\author{\IEEEauthorblockN{Zhihao Zhao}
\IEEEauthorblockA{\textit{Technical University of Munich }\\
Munich, Germany \\
zhihao.zhao@tum.de}
\and
\IEEEauthorblockN{Yinzheng Zhao}
\IEEEauthorblockA{\textit{ Technical University of Munich }\\
Munich, Germany \\
yinzheng.zhao@tum.de}
\and
\IEEEauthorblockN{Junjie Yang}
\IEEEauthorblockA{\textit{ Technical University of Munich }\\
Munich, Germany \\
junjie.yang@tum.de}
\and
\IEEEauthorblockN{Xiangtong Yao}
\IEEEauthorblockA{\textit{ Technical University of Munich }\\
Munich, Germany  \\
xiangtong.yao@tum.de}
\and
\IEEEauthorblockN{Quanmin Liang}
\IEEEauthorblockA{\textit{ Sun Yat-Sen University }\\
Guangzhou, China  \\
liangqm5@mail2.sysu.edu.cn}
\and
\IEEEauthorblockN{Daniel Zapp}
\IEEEauthorblockA{\textit{ Technical University of Munich }\\
Munich, Germany  \\
daniel.zapp@mri.tum.de}
\and
\IEEEauthorblockN{Kai Huang}
\IEEEauthorblockA{\textit{ Sun Yat-Sen University }\\
Guangzhou, China \\
huangk36@mail.sysu.edu.cn}
\and
\IEEEauthorblockN{Nassir Navab}
\IEEEauthorblockA{\textit {Technical University of Munich }\\
Munich, Germany \\
nassir.navab@tum.de}
\and
\IEEEauthorblockN{M.Ali Nasseri}
\IEEEauthorblockA{\textit{Technical University of Munich}\\
Munich, Germany \\
ali.nasseri@mri.tum.de}
}

\maketitle

\newcommand{\UOPSL}{{\textbf{UOPSL}}\xspace}
\begin{abstract}
Significant advancements in AI-driven multimodal medical image diagnosis have led to substantial improvements in ophthalmic disease identification in recent years. However, acquiring paired multimodal ophthalmic images remains prohibitively expensive. While fundus photography is simple and cost-effective, the limited availability of OCT data and inherent modality imbalance hinder further progress. Conventional approaches that rely solely on fundus or textual features often fail to capture fine-grained spatial information, as each imaging modality provides distinct cues about lesion predilection sites. In this study, we propose a novel unpaired multimodal framework \UOPSL that utilizes extensive OCT-derived spatial priors to dynamically identify predilection sites, enhancing fundus image-based disease recognition. Our approach bridges unpaired fundus and OCTs via extended disease text descriptions. Initially, we employ contrastive learning on a large corpus of unpaired OCT and fundus images while simultaneously learning the predilection sites matrix in the OCT latent space. Through extensive optimization, this matrix captures lesion localization patterns within the OCT feature space.
During the fine-tuning or inference phase of the downstream classification task based solely on fundus images, where paired OCT data is unavailable, we eliminate OCT input and utilize the predilection sites matrix to assist in fundus image classification learning. Extensive experiments conducted on 9 diverse datasets across 28 critical categories demonstrate that our framework outperforms existing benchmarks.
\end{abstract}


\section{Introduction}
Deep learning is rapidly advancing in the field of multimodal ophthalmic image diagnosis, as different imaging modalities provide distinct information about the predilection sites of various diseases \cite{yoo2019possibility,tong2020application}. However, the cost of acquiring different types of ophthalmic images varies significantly. For instance, fundus photography is a relatively inexpensive and straightforward imaging technique, whereas optical coherence tomography (OCT) is more costly \cite{yannuzzi2004ophthalmic}. As a result, obtaining \textbf{paired multimodal images is often challenging, leading to an imbalance in the availability of data across modalities}.
These challenges not only constrain the performance of existing AI methods but also limit their ability to fully leverage the rich, cross-modal disease cues.

Current retinal diagnostic models mainly focus on single-modal accuracy and the challenges of multimodal image fusion \cite{qiu2023visionfm}. Single-modality approaches often plateau in performance due to limited disease information from one device, underscoring the need for multimodal fusion. However, current multimodal approaches rely on paired data \cite{baltruvsaitis2018multimodal}, which is impractical given the scarcity and cost of OCT imaging. Models like CLIP \cite{conde2021clip} show promise in bridging visual and textual data but lack precise spatial alignment between text and disease-specific predilection sites. 
\textbf{OCT can effectively demonstrate the predilection sites of retinal diseases and provide detailed morphological and spatial localization information about lesions}, which cannot be fully captured through textual descriptions.

To address these challenges, we propose a novel unpaired multimodal framework \UOPSL that leverages extensive unpaired OCT-derived spatial priors to implicitly and dynamically learn the predilection sites of retinal diseases in the OCT latent space, thereby enhancing the diagnosis of diseases from fundus images. We initially collected a dataset comprising 83,687 high-quality fundus images and 80,923 OCT B-scans, which are weakly paired through 28 categories of retinal disease labels, along with expert-annotated disease descriptions. During the model pretraining phase, our approach first establishes associations between unpaired fundus images and OCT scans by utilizing extended disease text descriptions as an intermediate bridge. By harnessing the capabilities of CLIP \cite{conde2021clip}, we project heterogeneous data sources into a unified feature space, conducting contrastive learning between a large number of fundus images, text descriptions encapsulating disease semantics, and unpaired OCT images that provide precise disease localization. Simultaneously, for the OCT latent space, we dynamically learn an additional disease-specific predilection matrix, which is element-wise multiplied with the latent features of OCT to capture information about disease predilection sites within the OCT latent space. Subsequently, we employ cross-attention to associate the OCT-derived predilection sites and expert-annotated text descriptions. Finally, during downstream classification or inference, in the absence of paired OCT data, we exclude the OCT input and rely solely on the learned lesion predilection sites to enhance the classification of fundus images.

The main contributions of this work are as follows. First, we constructed a high-quality dataset consisting of fundus photographs and OCT B-scan images, weakly paired through 28 retinal disease categories and complemented by expert-annotated textual descriptions. Second, we proposed a novel unpaired multimodal framework that dynamically learns disease-specific predilection sites from OCT-derived spatial priors, thereby enhancing disease recognition in fundus images without requiring paired OCT data during inference. Finally, we mathematically validated our approach and conducted extensive experiments across nine diverse datasets and 28 segmentation categories, consistently demonstrating superior performance compared to existing methods.

\section{Related Work}
\subsection{Unpaired Multimodal Ophthalmic Image Analysis }

The most commonly used imaging modalities in ophthalmology are Color Fundus Photography (CFP) and Optical Coherence Tomography (OCT). To improve disease diagnosis, AI models are increasingly integrating the advantages of both modalities \cite{he2021multi}. Much research has focused on fusing information from these different modalities, typically by using separate encoder architectures to extract CFP and OCT features, followed by various feature fusion methods \cite{liu2024cross, tian2024tagat, kang2021multimodal}. However, acquiring sufficient training data for specific imaging modalities is often challenging in practice \cite{nakada2023understanding}. To address this, MultiEYE \cite{wang2024multieye} proposed an OCT-assisted Concept Distillation method (OCT-CoDA). This method extracts disease-related knowledge from OCT images using semantically rich concepts and transfers it to a CFP model. The framework is designed to leverage spatial priors from unpaired OCT data, thereby eliminating the need for explicit CFP-OCT pairs.

\subsection{Foundational Models in Ophthalmology}

A foundation model refers to a large-scale artificial intelligence model trained on extensive and varied data, enabling its adaptation to a wide array of downstream tasks. In 2023, Zhou et al. released RETFound, a foundation model specifically designed for retinal images. The training of RETFound \cite{zhou2023foundation} involved a generative self-supervised method called Masked Autoencoder (MAE), first applied to 1.3 million natural images and subsequently to 1.6 million retinal images. Following a similar self-supervised paradigm, the UrFound model \cite{yu2024urfound} processes simultaneous inputs of Optical Coherence Tomography (OCT) and Color Fundus Photography (CFP), incorporates domain knowledge into its representation learning, and encodes expert annotations via textual supervision.
In addition to these self-supervised approaches, CLIP-based vision-language models are gaining prominence. FLAIR \cite{silva2025foundation}, for example, integrates expert domain knowledge in the form of descriptive text prompts to bolster classification supervision where labels are less informative. Another model, Ret-CLIP \cite{du2024ret}, extracts general-purpose features from Color Fundus Photographs (CFP) and utilizes a triplet optimization strategy—focusing on left-eye, right-eye, and patient levels—to mirror authentic clinical scenarios. ViLReF \cite{yang2024vilref} assists label extraction with expert knowledge and proposes a new constraint, the Weighted Similarity Coupling Loss, for dynamically adjusting the rate at which sample pairs are separated within the feature space. Meanwhile, RetiZero \cite{wang2025enhancing} consolidates knowledge of over 400 fundus diseases from public datasets, ophthalmic literature, and online sources, encompassing a wide range of ethnic and national origins.
The integration of Large Language Models (LLMs) into the ophthalmic domain is also a growing trend. Vision-Language Models (VLMs) such as VisionUnite \cite{li2024visionunite} have emerged as novel ophthalmic foundation models grounded in clinical knowledge; they are pre-trained on massive datasets and then fine-tuned on the specialized MMFundus dataset. Similarly, LEME \cite{gilson2024language} was first pre-trained on the Llama2 70B architecture and then further refined through fine-tuning on a large corpus of authentic materials, including specialized ophthalmic case reports, abstracts, and open-source learning resources.

\subsection{Retina Disease Predilection Sites and Spatial Priors}

It is well-established that human diseases exhibit specific sites of predilection, a principle that also applies to the retina \cite{ascher1952there,herwig2018feature}. For instance, macular edema preferentially occurs in the macular region, which is a known characteristic of the disease. Due to its ability to resolve subtle morphological changes, Optical Coherence Tomography (OCT) imaging plays a crucial role in localizing the predilection sites of retinal diseases \cite{vogl2015spatio,nam2024spatial}. Likewise, the anatomical location of organs and tissues can provide valuable prior knowledge to aid a model in disease diagnosis \cite{fazekas2022sd}. When such prior knowledge about disease location is available, we can leverage attention mechanisms \cite{vaswani2017attention} within the current pipeline to guide the model's focus toward these predilection regions.

\section{Methodology}
\label{sec:methodology}

\begin{figure*}[ht] \centering
	\centering
	\includegraphics[width=0.95\textwidth]{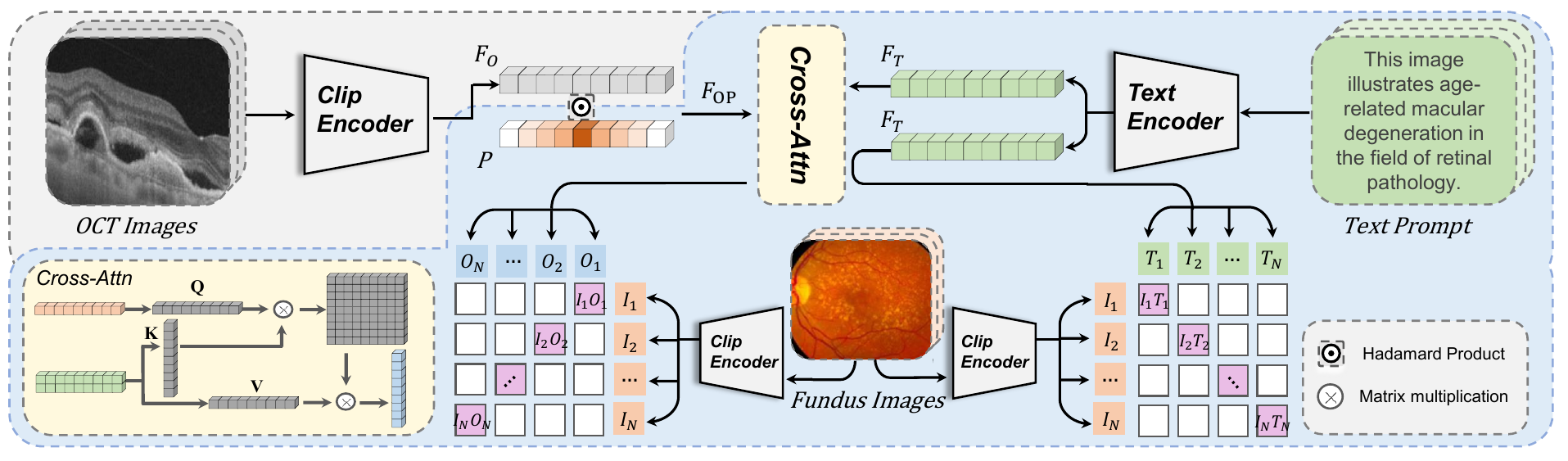}
	\caption{Overview of our proposed \UOPSL. During the model training phase, it concurrently accepts OCT, fundus, and disease text as inputs while dynamically learning the predilection sites matrix $P$. For fine-tuning in downstream classification tasks or the inference phase, the OCT module can optionally be excluded from the input.} 
	\label{fig:flowchart}
\end{figure*}

In Fig.~\ref{fig:flowchart}, we provide a comprehensive delineation of the methodology we have proposed. Our model \UOPSL is designed to receive three types of inputs during the training phase, which include fundus images, OCT images, and textual descriptions of diseases. 
In our training phase, fundus images, OCT images, and extended disease category texts with the same disease are jointly processed. Specifically, fundus images are encoded into feature space $F_{F}$, OCT images into $F_{O}$, and texts into $F_{T}$. A learnable predilection sites matrix $P$ is used to weight the OCT features, which are element-wise multiplied with the latent features of OCT to capture information about disease predilection sites within the OCT latent space. Then the weighted features and the $F_{T}$ are fed into cross-attention.
Cross-attention enables the association between textual descriptions of diseases and disease-specific predilection sites in the OCT latent space. In this way, expert labels can bridge the connection between fundus images and unpaired OCT data, linking them through disease-specific predilection sites.
After extensive data training, the predilection sites matrix $P$ becomes adept at indicating the precise locations of diseases within the OCT latent space. During the inference phase, the requirement for OCT image input is obviated; instead, the predilection sites matrix $P$ is utilized to effectively locate the corresponding disease positions within the OCT space from the textual latent space. This innovative approach leverages the textual latent space as an intermediary conduit, thereby enabling the effective utilization of predilection sites information from the unpaired OCT latent space when computing disease features in fundus images.

\subsection{Latent Space Encoder }
Let $I_F$ denote a fundus image, $I_O$ an OCT image, and $T_{\text{raw}}$ the raw text label corresponding to a disease category. 
The fundus images and OCT B-scans with the same disease are encoded to the embedding dimension $d$ using the ViT-based vision encoder $E_{CLIP}$ from the MetaCLIP \cite{xu2023demystifying} ViT-B/16 model. The expert annotation text prompt is embedded using BioMed-BERT \cite{chakraborty2020biomedbert} text encoder $E_T$ 
\begin{equation}
\begin{aligned}
  F_{F} &= E_{\text{CLIP}}(I_{F}) \in \mathbb{R}^{d}, \\
  F_{O} &= E_{\text{CLIP}}(I_{O}) \in \mathbb{R}^{d}, \\
  F_{T} &= E_t(T_{\text{raw}}) \in \mathbb{R}^{d}
\end{aligned}
\end{equation}
  
\subsection{Learnable OCT-Conditioned Predilection Matrix and Cross-Attention Mechanism}

\subsubsection{Learnable OCT-Conditioned Predilection Matrix. }
To enable the disease label-based text latent space to capture the disease predilection sites information from the OCT latent space, we employed a predilection sites matrix $P$ that is optimized during the training process, thereby allowing it to capture latent spatial importance cues. We first compute an element-wise production (Hadamard product at the pixel level) between $F_{O}$ and $P$. Subsequently, the OCT latent space and the text latent space are interconnected through cross-attention.
\begin{equation}
    F_{OP} = F_{O} \odot sigmoid(P),\quad P \in \mathbb{R}^{d}.
\end{equation}
where $\odot$ denotes element-wise product, revealing predilection sites in the OCT latent space.

\subsubsection{Cross-Attention Mechanism on OCT and Text Latent Space. } The reweighted OCT feature $F_{OP}$ is then fused with the text representation $F_{T}$ using a cross-attention mechanism to produce a refined spatial cue $F_{OT}$ that encapsulates both the OCT-derived spatial information and the semantic guidance from the disease text.

First, we compute query, key, and value projections:
\begin{equation}
\begin{aligned}
    Q &= F_{OP} W_q \in \mathbb{R}^{d}, \\
    K &= F_{T} W_k \in \mathbb{R}^{d}, \\
    V &= F_{T} W_v \in \mathbb{R}^{d}
\end{aligned}
\end{equation}
where $W_q$, $W_k$, and $W_v \in \mathbb{R}^{d \times d}$ are learnable weight matrices.

Next, we calculate the attention weights $Att$ and then obtain the refined spatial cue $F_{OT}$:
\begin{equation}
Att = \operatorname{softmax}\left(\frac{Q K^\top}{\sqrt{d}}\right) \in \mathbb{R}, \quad F_{OT} = Att \, V \in \mathbb{R}^{d}.
\end{equation}
This process establishes a robust linkage between the predilection sites matrix of the OCT space and the textual feature space of diseases, enabling the direct utilization of OCT predilection sites information during the inference phase.

\subsection{Contrastive Learning Objectives}
To align the different modalities in the latent space, we employ contrastive learning in two branches:
\subsubsection{Fundus and OCT  Contrast: }  
We encourage the fundus representation $F$ to be aligned with the refined spatial cue $F_{OP}$ via a CLIP-like contrastive loss:
  \begin{equation}
  \mathcal{L}_{F_F,F_{OP}} = -\log \frac{\exp(\operatorname{sim}(F_F, F_{OP})/\tau)}{\sum_{F_{OP}'} \exp(\operatorname{sim}(F_F, F_{OP}')/\tau)},
  \end{equation}
  where $\operatorname{sim}(\cdot,\cdot)$ denotes cosine similarity and $\tau$ is a temperature parameter.

\subsubsection{Fundus and Text Contrast: }  
Similarly, a contrastive loss aligns the fundus features $F_F$ directly with the text embedding $F_T$:
  \begin{equation}
  \mathcal{L}_{F_F,F_T} = -\log \frac{\exp(\operatorname{sim}(F_F, F_T)/\tau)}{\sum_{F_T'} \exp(\operatorname{sim}(F_F, F_T')/\tau)}.
  \end{equation}

The overall training loss is a weighted sum:
\begin{equation}
\mathcal{L} = \lambda_1 \mathcal{L}_{F_F,F_{OP}} + \lambda_2 \mathcal{L}_{F_F,F_T},
\end{equation}
We empirically set hyperparameters $\lambda_1$ and $\lambda_2$ to 0.4 and 0.6, respectively.

\subsection{Fine-tune and Inference Phase}
\begin{figure}[ht] \centering
	\centering
	\includegraphics[width=0.4\textwidth]{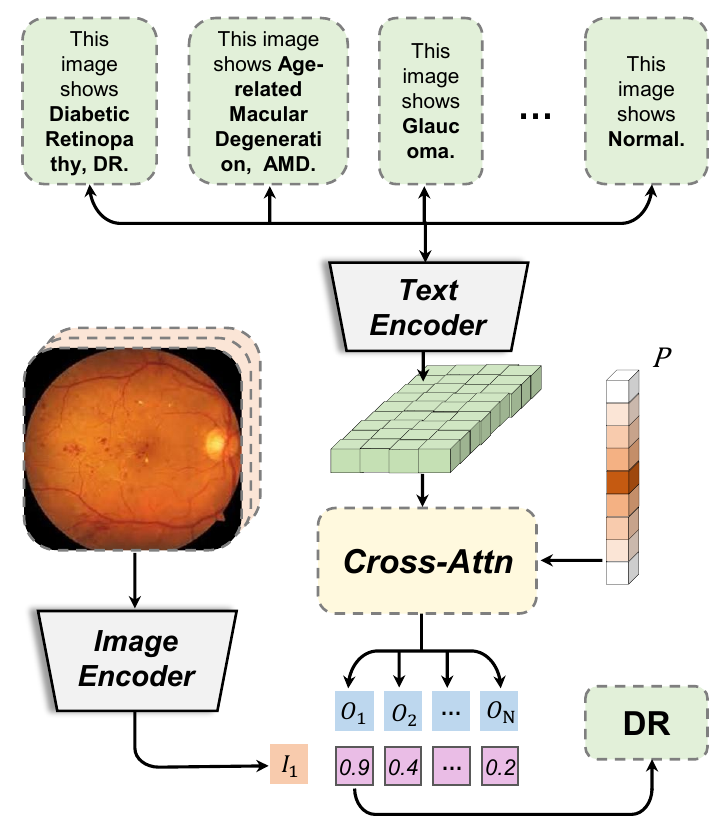}
	\caption{Overview of the fine-tuning and inference phases.} 
	\label{fig:finetuneinference}
\end{figure}

During the fine-tuning and inference phase, our model operates without the OCT (Optical Coherence Tomography) encoder. To compensate for the absence of the OCT-derived feature space, we introduce a pre-trained predilection matrix $P$. This matrix serves as a proxy, encoding the spatial priors of typical disease predilection locations.
As shown in Fig.~\ref{fig:finetuneinference}, the process begins by converting a set of disease labels into descriptive textual prompts using a predefined template, such as "This retina fundus image shows \{Diabetic Retinopathy\}, \{DR\}". These prompts are then processed by a text encoder to generate an initial textual feature matrix.
Subsequently, this textual feature matrix is fused with the predilection matrix P via a Cross-Attention mechanism. The objective of this operation is to spatially accentuate the regions corresponding to the predilection sites of various ophthalmic diseases. As a result, the refined latent matrix exhibits significantly higher activation values at these key pathological locations, while values in other regions are suppressed.
Subsequently, a similarity matrix is computed between the text and image embedding spaces. As the text features have been tailored to highlight vulnerable areas, the similarities calculated for these positions are consequently amplified. The text that exhibits the highest similarity to the image via this procedure serves as our label.

\subsection{Proof of Consistency in Spatial Importance Ranking without OCT Images}

During the fine-tuning of the downstream task for disease classification using single-modal fundus images, as well as in the inference phase, since paired OCT images are unavailable,  only the fundus image $I_F$ and the disease text $T_{\text{raw}}$ are available. Therefore, our objective is to develop a method for identifying disease predilection sites within the OCT space without the need for paired data. This means we only require the positional representation of disease information within the OCT latent space. Consequently, the input can be an abstract, conceptual image that amalgamates information from various diseases, which in turn is highly correlated with the predilection sites matrix $P$.

This analysis theoretically justifies substituting a pre-learned disease predilection matrix, denoted by $P$, for the latent features of OCT images, $F_O$, within a cross-attention mechanism. This scenario arises during fine-tuning and inference phases where paired OCT images are unavailable. We demonstrate that, under the assumption of high linear correlation between $F_O$ and $P$, using $P$ as a proxy can produce a consistent attention-based importance ranking.

\subsection{Problem Formulation}

Let the latent feature matrix from an OCT image be $F_O \in \mathbb{R}^{n \times d}$ and the learned spatial prior matrix be $\hat{P} \in \mathbb{R}^{n \times d}$ (normalized via sigmoid, with elements in $[0, 1]$). The text feature is denoted $F_T$.

Our central hypothesis is that when the model is sufficiently well-trained, the information encoded in the OCT feature map $F_O$ is highly correlated with the spatial disease priors in $\hat{P}$. 
That is, under ideal, well-trained conditions, $F_O = \alpha \hat{P}$. 
Next, we need to analyze the reasonableness of removing the OCT image when the model is sufficiently well-trained. 
However, we know that $F_O$ and $\hat{P}$ cannot be perfectly linearly correlated. 
Therefore, we introduce a small random perturbation $B$. 
After relaxing the condition, we need to prove that when $F_O = \alpha \hat{P} + B$, where $B$ is a small random perturbation, it is still reasonable for our model to remove the OCT image.

\subsection{Comparison of Attention Formulations}

We compare an ideal (but infeasible) scenario that uses OCT features with the practical (feasible) scenario that uses the proxy matrix $\hat{P}$.

Scenario 1: Ideal Attention with Prior-Modulated OCT Features (Infeasible)
In an ideal setting, the most effective query would not use the raw feature $F_O$, but rather a feature map that is spatially modulated by the priors themselves. We define this ideal feature source as $F_{OP} = F_O \odot \hat{P}$, where $\odot$ is the element-wise Hadamard product.

The query matrix $Q^{(OP)}$ is thus:
\begin{equation}
    Q^{(OP)} = (F_O \odot \hat{P}) W_Q
\end{equation}
The key vector $K$ is derived from the text features, $K = F_T W_K$. The resulting attention $a^{(OP)}$ for each of the $n$ spatial locations is
\begin{equation}
    a^{(OP)}_i = ((F_O \odot \hat{P})_i W_Q) \cdot (F_T W_K) \quad \text{for } i=1, \dots, n.
    \label{equ:9}
\end{equation}

Scenario 2: Proposed Attention with Proxy Matrix (Feasible)
In practice, without $F_O$, we use $\hat{P}$ directly to form the query. For each spatial location $i$, the query vector is formed from the $i$-th row of $\hat{P}$.

The query matrix $Q^{(P)}$ is:
\begin{equation}
    Q^{(P)} = \hat{P} W_Q
\end{equation}
The practical attention  $a^{(P)}$ for each spatial location $i$ is:
\begin{equation}
    a^{(P)}_i = (\hat{P}_i W_Q) \cdot (F_T W_K) \quad \text{for } i=1, \dots, n.
\end{equation}

\subsection{Analysis of Rank Equivalence}

To analyze the relationship, we substitute our hypothesis  into the ideal query source, $F_{OP}$:
\begin{equation}
\begin{aligned}
    F_{OP} &= (\alpha \hat{P} + B) \odot \hat{P} \\
           &= \alpha (\hat{P} \odot \hat{P}) + (B \odot \hat{P}) \\
           &= \alpha \hat{P}^2 + B'
\end{aligned}
\end{equation}
Here, $\hat{P}^2$ denotes the element-wise square of the matrix $\hat{P}$, and $B' = B \odot \hat{P}$ is a new, still low-magnitude, noise term.
Substituting this into the ideal logit Equation~\ref{equ:9} for each spatial location $i$:
\begin{equation}
\begin{aligned}
    a^{(OP)}_i &= ((\alpha \hat{P}^2 + B')_i W_Q) \cdot (F_T W_K) \\
               &\approx \alpha ((\hat{P}_i^2 W_Q) \cdot (F_T W_K)) \quad \text{for } i=1, \dots, n.
\end{aligned}
\end{equation}
Finally, we compare the approximate ideal attention logit with the practical attention logit for each spatial location $i$. 
\begin{equation}
\begin{aligned}
      a^{(OP)}_i &\approx \alpha \cdot \left( (\hat{P}_i^2 W_Q) \cdot  (F_T W_K) \right) \quad  \\
      \quad a^{(P)}_i &= \left( (\hat{P}_i W_Q) \cdot  (F_T W_K) \right) \quad \text{for } i=1, \dots, n.  
\end{aligned}
\end{equation}

The attention logits  are not linearly proportional due to the presence of $\hat{P}^2$ in the ideal case and $\hat{P}$ in the practical case. However, our goal is to preserve the ranking of spatial importance.

Let $z_i$ be the vector of pre-projection scores for the $i$-th spatial location, such that the final attention logit $a_i$ is a function of $z_i$. In our case, $z_i^{(OP)} \propto \hat{P}_i^2$ and $z_i^{(P)} \propto \hat{P}_i$. Since all elements of $\hat{P}$ are non-negative (output of a sigmoid), the squaring function $f(x)=x^2$ is strictly monotonically increasing for non-negative inputs. This means that for any two spatial locations $j$ and $k$:
\begin{equation}
    \text{if } \hat{P}_j > \hat{P}_k, \text{ then } \hat{P}_j^2 > \hat{P}_k^2
\end{equation}
Therefore, the rank ordering of the elements is preserved across all $n$ locations:
\begin{equation}
    \text{rank}(a^{(OP)}) = \text{rank}(a^{(P)})
\end{equation}

\subsection{Conclusion}
Based on our analysis, substituting $\hat{P}$ for the unavailable OCT features is a principled approach, particularly after the model has been sufficiently trained on the distributional space of the OCT data.
While the resulting attention scores from `softmax` may differ in magnitude, their relative ranking is preserved for all $n$ spatial locations. This ensures that the attention mechanism correctly identifies the same hierarchy of disease-relevant spatial locations.

\section{Experiments}
\newsavebox\CBox
\def\textBF#1{\sbox\CBox{#1}\resizebox{\wd\CBox}{\ht\CBox}{\textbf{#1}}}

\subsection{Setup Details}

\subsubsection{Data collection and processing }

We combined 25 public fundus datasets \cite{silva2025foundation} and two private hospital datasets, using the EyeQ \cite{fu2019evaluation} to filter out low-quality images. We then filtered the CFP and OCT data to include only the following categories: Normal, Glaucoma (2 levels), Cataract, DR (4 levels), AMD, DME, CNV, RAO, and RVO, resulting in 83,687 fundus images. We also gathered 80,923 OCT B-scans from public sources. Since most public ophthalmology datasets provide classification labels rather than natural language text, in our study, we adopted the FLAIR approach \cite{silva2025foundation}, which involves generating ten distinct descriptions for each disease. During the training phase, a single description was randomly selected for each instance. For example, to describe "hard exudates," we used various expressions such as "This image has \{Hard Exudates\}, with small, well-defined yellowish-white deposits." This method of using diverse, randomly selected descriptions during training was employed to enhance data diversity and improve the model's generalization capabilities.

\subsubsection{Testing Datasets for Transferability}
\label{sec:datafortrans}
To evaluate our model's capacity to transfer learned representations to downstream tasks.
In our pre-training phase, we deliberately excluded data from the following datasets to facilitate algorithm evaluation: Messidor \cite{decenciere2014feedback}, IDRID \cite{porwal2020idrid}, APTOS-2019\cite{aptos}, PAPILA \cite{kovalyk2022papila}, Glaucoma Fundus (GF) \cite{ahn2018deep}, JSIEC \cite{cen2021automatic}, Retina  \footnote{https://www.kaggle.com/datasets/jr2ngb/cataractdataset}, RFMID \cite{pachade2021retinal}, and ODIR \cite{challenge2019peking}. Regarding the partitioning of datasets into training, testing, and validation sets, we adhered to the official division standards for datasets like IDRID \cite{porwal2020idrid} and RFMID \cite{pachade2021retinal}. For the remaining test datasets, we followed the ratio delineated in Ret-CLIP \cite{du2024ret}, allocating 56\% for training, 14\% for validation, and 30\% for testing.

\subsubsection{Testing Datasets for Domain Shift}

To evaluate performance under domain shift, the experimental setup required a constant label set while the image data exhibited a domain shift. Consequently, from the 9 datasets excluded from the training set as described in Section ~\ref{sec:datafortrans}, we selected IDRID, APTOS-2019, PAPILA, and Retina for testing. These datasets were chosen because their labels are fully encompassed by the training set, yet they originate from different image domains.
\subsubsection{Testing Datasets for Unseen Categories}
To evaluate the zero-shot generalization of our method on unseen categories, we designated four diseases as unseen categories: retinitis pigmentosa (RP), macular hole (MH), hypertension (HT), and myopia (MYA). All samples corresponding to these conditions were completely excluded from the training process. Subsequently, two dedicated evaluation subsets were constructed:
The JEIEC114 dataset, sourced from the JEIEC collection, is comprised of 114 images distributed across four distinct ophthalmological categories. These categories are RP with 22 images, MH with 15 images, HT with 23 images, and MYA with 54 images.
The ODIR360 dataset, a subset of the ODIR-5K collection, consists of 360 images classified into two categories: HT, which includes 128 images, and MYA, with the remaining 232 images.
The composition and categories of the datasets for Domain Shift and Unseen Categories are detailed in TABLE ~\ref{tab:dataset}.

\begin{table}[h!]
\centering
\caption{Dataset distribution for evaluating the generalization of foundation model}
\label{tab:dataset}
\scriptsize
\begin{tblr}{
  width = 0.95\linewidth, 
  colspec = {Q[l, m] Q[l, m] Q[c, m] X[l, m]}, 
  cell{2}{1} = {r=4}{}, 
  cell{6}{1} = {r=2}{},
  hline{1,8} = {1-4}{1.2pt}, 
  hline{2,6} = {0.8pt}, 
  hline{3-4,5,7} = {dashed, 0.4pt}, 
  vline{2,3} = {0.8pt}, 
}
\textbf{Scenario}  &\textbf{Dataset} & \textbf{Images}  & \textbf{Labels} \\

\SetCell[c=1]{l, font=\itshape} Domain shift & IDRID  & 516  & noDR, mildDR, modDR, sevDR, prolDR. \\
& APTOS2019  & 3662 & N, mildDR, modDR, sevDR, prolDR. \\
& PAPILA     & 488  & N, Early G, Advanced G. \\
& Retina    & 601 & N, G, Cataract, Other\\
\SetCell[c=1]{l, font=\itshape} Unseen categories & JSIEC 22+15+23+54      & 114   & RP,HT,MH,MYA \\
& ODIR 128+232 & 360  & HT, MYA \\

\end{tblr}
\end{table}

\subsubsection{Data preprocessing and augmentation}
CFP and OCT images are padded to a 1:1 ratio, resized to 512×512, and augmented during training with random cropping (resized to 224×224 for ViT), horizontal flipping, color jitter, and normalization. For text, up to 10 ChatGPT-generated descriptions per disease label are used, with dynamic pairing across epochs.
\subsubsection{Experimental Settings}
The model was pre-trained using PyTorch on four Nvidia A5000 24GB GPUs, with a total training duration of 200 epochs, including a 50-epoch warm-up period, and a learning rate set at 5e-5. During the fine-tuning phase, we implemented the Linear Probing approach, freezing both the visual and textual encoders. The learning rate was adjusted to 1e-5, with a fine-tuning duration of 50 epochs and a 10-epoch warm-up phase.

\subsubsection{Evaluation Metric}
To illustrate the advantages of our approach, we conducted a comparative analysis of our models against Unimed-CLIP \cite{uzair2024unimed}, RETFound \cite{zhou2023foundation}, FLAIR \cite{silva2025foundation}, UrFound \cite{yu2024urfound}, and RET-CLIP \cite{du2024ret}. In addition to multimodal foundation models, we included single-modality methods for comparison. Specifically, we used the ViT-B/16 model to serve as the vision encoder in our proposed model as a standalone single-modality baseline. To ensure a fair and comprehensive comparison, we performed independent supervised fine-tuning (SFT) of this ViT model on each individual dataset.
We used the area under the receiver operating characteristic curve (ROC) and the area under the precision-recall curve (PRC) as evaluation metrics, averaging results over five runs with different random seeds for each downstream dataset.

\subsection{Evaluation and Results}

\subsubsection{Transferability}

\textbf{For linear probing}.  We initially conducted pre-training using the proposed model, followed by freezing the image and text encoders of the model. Concurrently, we removed the OCT encoder component from the model.  Subsequently, we performed linear probing to fine-tune the classifier for each retinal disease diagnosis task. We feed the features from the vision encoder into a linear classifier. This classifier's parameters are then fine-tuned using the same multi-class logistic regression optimizer as described in the CLIP \cite{radford2021learning} paper. The specific quantitative results are presented in TABLE ~\ref{tab:1} and ~\ref{tab:2}. TABLE ~\ref{tab:1} encompasses two diseases: diabetic retinopathy (4 classes) and glaucoma (2 classes). The results indicate that the model's performance on the glaucoma dataset is relatively similar across different models, whereas our model consistently outperforms others on the DR dataset. This superiority can be attributed to our model's enhanced capability to capture features of subtle lesion areas in the context of complex disease characteristics. For the glaucoma dataset, the regions of interest in fundus images are predominantly determined by the optic cup and disc areas, which are distinctly noticeable and thus equally attended to by different models.

The superior capability of our model in capturing complex lesion features within images is further substantiated in TABLE ~\ref{tab:2}, which delineates the multi-disease classification tasks for retinal fundus images. In these tasks, our model consistently surpasses other models, with its PRC (Precision-Recall Curve) metric notably exceeding those of its counterparts. This underscores the model's enhanced performance advantage, particularly in scenarios characterized by class imbalance.

\begin{table}[h]
\centering
\caption{Quantitative Analysis Results on DR and  Glaucoma Datasets}
\label{tab:1}
\resizebox{.98\linewidth}{!}{
\begin{tblr}{
  cell{1}{1} = {r=3}{},
  cell{1}{2} = {c=6}{c},
  cell{1}{8} = {c=4}{c},
  cell{2}{2} = {c=2}{c},
  cell{2}{4} = {c=2}{c},
  cell{2}{6} = {c=2}{c},
  cell{2}{8} = {c=2}{c},
  cell{2}{10} = {c=2}{c},
  vline{2,8} = {1}{},
  vline{2,4,6,8,10} = {2-10}{},
  hline{1,4,10,11} = {-}{},
  hline{2} = {2-11}{},
}
Models      & DR    &       &           &       &          &       & Glaucoma &       &                &                \\
            & IDRID &       & APTOS2019 &       & Messidor &       & PAPILA   &       & GF             &                \\
            & ROC   & PRC   & ROC       & PRC   & ROC      & PRC   & ROC      & PRC   & ROC            & PRC            \\
Unimed-CLIP  \cite{uzair2024unimed} & 0.633 & 0.336 & 0.806     & 0.429 & 0.842    & 0.488 & 0.658    & 0.473 & 0.863          & 0.716          \\
RET-Found \cite{zhou2023foundation}   & 0.665 & 0.368 & 0.745     & 0.370 & 0.864    & 0.586 & 0.620    & 0.511 & 0.899          & 0.773          \\
FLAIR \cite{silva2025foundation}      & 0.700 & 0.475 & 0.849     & 0.515 & 0.819    & 0.483 & 0.746    & 0.595 & 0.872          & 0.672          \\
UrFound \cite{yu2024urfound}    & 0.852 & 0.577 & 0.949     & 0.716 & 0.882    & 0.608 & 0.783    & 0.625 & 0.958          & 0.880          \\
RET-CLIP \cite{du2024ret}   & 0.856 & 0.616 & 0.923     & 0.656 & \textBF{0.898}    & 0.609 & 0.775    & 0.667 & 0.893          & 0.789          \\
\textbf{ViT}$_{SFT}$ \cite{dosovitskiy2020image} & 0.852 & 0.607 & 0.937     & 0.658 & 0.884    & 0.617 & 0.747    & 0.602 & 0.927          & 0.831          \\
$\UOPSL_{LP}$ (ours)        & \textBF{0.865} & \textBF{0.633} & \textBF{0.960}     & \textBF{0.738} & 0.890    & \textBF{0.621} & \textBF{0.818}    & \textBF{0.729} & \textBF{0.958}          & \textBF{0.890}  \\   
\end{tblr}
}
\end{table}

\begin{table}[h]
\centering
\caption{Quantitative Analysis Results on Multi-disease Datasets}
\label{tab:2}
\resizebox{.98\linewidth}{!}{
\begin{tblr}{
  cell{1}{1} = {r=2}{},
  cell{1}{2} = {c=2}{c},
  cell{1}{4} = {c=2}{c},
  cell{1}{6} = {c=2}{},
  cell{1}{8} = {c=2}{},
  vline{2,4,6,8} = {1-9}{},
  hline{1,3,9-10} = {-}{},
  hline{2} = {2-9}{},
}
Models      & JSIEC &       & Retina &       & RFMID &       & ODIR  &       \\
            & ROC   & PRC   & ROC    & PRC   & ROC   & PRC   & ROC   & PRC   \\
Unimed-CLIP \cite{uzair2024unimed} & 0.783 & 0.239 & 0.738  & 0.514 & 0.819 & 0.293 & 0.801 & 0.483 \\
RET-Found \cite{zhou2023foundation}   & 0.704 & 0.167 & 0.630  & 0.434 & 0.842 & 0.409 & 0.738 & 0.401 \\
FLAIR \cite{silva2025foundation}      & 0.843 & 0.304 & 0.773  & 0.557 & 0.773 & 0.254 & 0.858 & 0.531 \\
UrFound \cite{yu2024urfound}    & 0.995 & 0.923 & 0.901  & 0.793 & 0.912 & 0.538 & 0.909 & 0.679 \\
RET-CLIP \cite{du2024ret}   & 0.982 & 0.855 & 0.935  & 0.864 & 0.925 & 0.552 & 0.902 & 0.682 \\
\textbf{ViT}$_{SFT}$ \cite{dosovitskiy2020image} & 0.983 & \textBF{0.937} & 0.906  & 0.804 & 0.908 & 0.559 & 0.913 & 0.714 \\
$\UOPSL_{LP}$ (ours)      & \textBF{0.997} & \textBF{0.932} & \textBF{0.945}  & \textBF{0.877} & \textBF{0.940} & \textBF{0.576} & \textBF{0.919} & \textBF{0.721}
\end{tblr}
}
\end{table}

\subsubsection{Generalization on Domain Shift}

To evaluate the generalization capability of the foundation model under a domain shift, we conducted a zero-shot inference without fine-tuning any trainable parameters. Specifically, we directly inputted a text prompt, constructed by integrating class names with predefined label templates, into the pretrained model for prediction.
Our comparative analysis in TABLE ~\ref{tab:domainshift} involved several CLIP-based zero-shot models and conventionally supervised models. The results indicate that our pretrained model substantially outperforms standard CLIP-based zero-shot approaches and surpasses the performance of traditionally supervised models.
Furthermore, we extended our comparison to include linear probing and full fine-tuning methodologies. On the Diabetic Retinopathy (DR) and Glaucoma detection tasks, our model's zero-shot performance was on par with that of linear probing.

\begin{table}[h]
\centering
\caption{ Generalization Capability on Domain Shift}
\label{tab:domainshift}
\resizebox{.98\linewidth}{!}{
\begin{tblr}{
  cell{1}{1} = {r=2}{},
  cell{1}{2} = {c=2}{c},
  cell{1}{4} = {c=2}{c},
  cell{1}{6} = {c=2}{},
  cell{1}{8} = {c=2}{},
  vline{2,4,6,8} = {1-8}{},
  hline{1,3,8-9} = {-}{},
  hline{2} = {2-9}{},
}
Models      & JSIEC &       & Retina &       & RFMID &       & ODIR  &       \\
            & ROC   & PRC   & ROC    & PRC   & ROC   & PRC   & ROC   & PRC   \\
Unimed-CLIP \cite{uzair2024unimed} & 0.783 & 0.239 & 0.738  & 0.514 & 0.819 & 0.293 & 0.801 & 0.483 \\
RET-Found \cite{zhou2023foundation}   & 0.704 & 0.167 & 0.630  & 0.434 & 0.842 & 0.409 & 0.738 & 0.401 \\
FLAIR \cite{silva2025foundation}      & 0.843 & 0.304 & 0.773  & 0.557 & 0.773 & 0.254 & 0.858 & 0.531 \\
UrFound \cite{yu2024urfound}    & 0.995 & 0.923 & 0.901  & 0.793 & 0.912 & 0.538 & 0.909 & 0.679 \\
RET-CLIP \cite{du2024ret}   & 0.982 & 0.855 & 0.935  & 0.864 & 0.925 & 0.552 & 0.902 & 0.682 \\
\UOPSL (ours)       & \textBF{0.997} & \textBF{0.972} & \textBF{0.945}  & \textBF{0.877} & \textBF{0.940} & \textBF{0.576} & \textBF{0.919} & \textBF{0.721}
\end{tblr}
}
\end{table}

\subsubsection{Generalization on Unseen Classes}
\begin{figure}[htbp]
    \centering

    \begin{subfigure}[b]{0.24\textwidth}
        \centering
        \includegraphics[width=\linewidth]{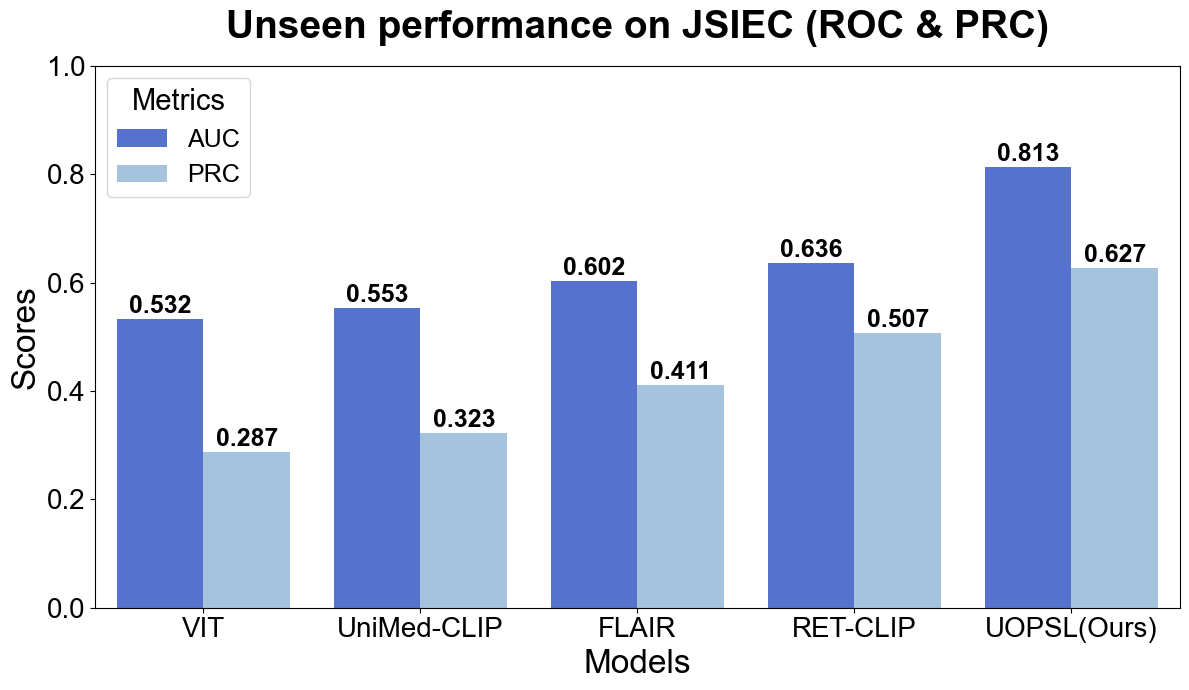}
        \label{fig:jsiec}
    \end{subfigure}
    \hfill 
    \begin{subfigure}[b]{0.24\textwidth}
        \centering
        \includegraphics[width=\linewidth]{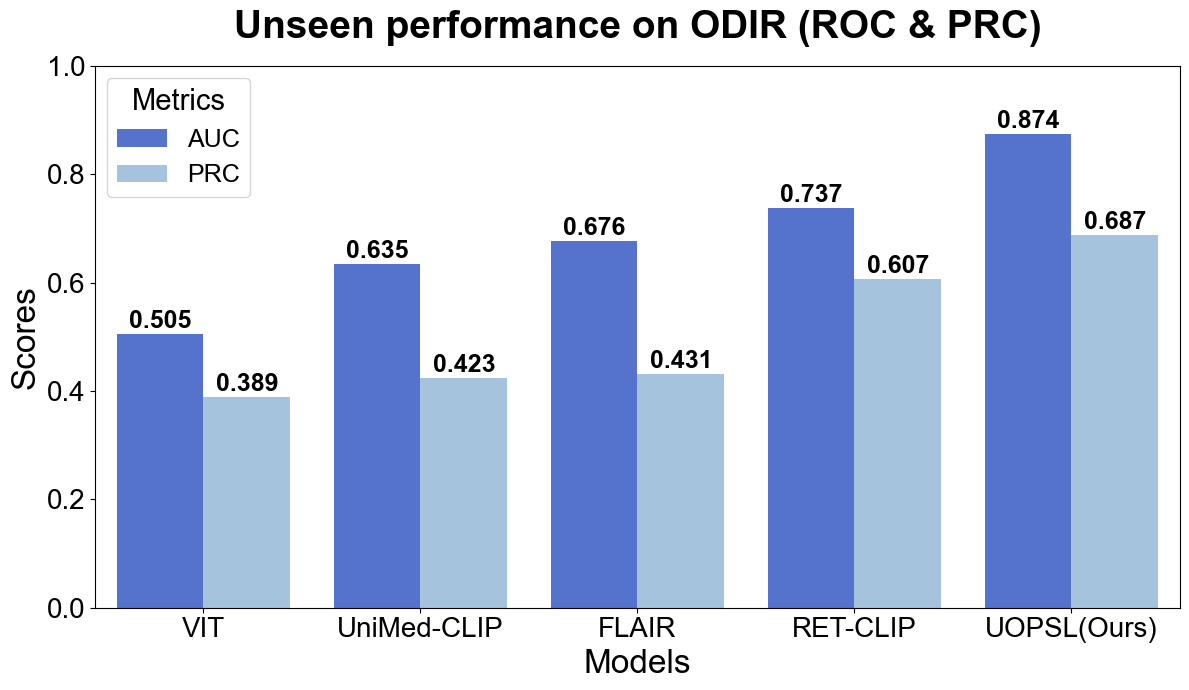}
        \label{fig:odir}
    \end{subfigure}

    \caption{ Generalization Capability on Unseen Classes}
    \label{fig:unseen}
\end{figure}

To evaluate  our model's zero-shot capability and to confirm that the predilection matrix effectively learns a comprehensive set of disease features within the OCT space, we conducted a specific analysis. This was based on the premise that even if a disease label has never been explicitly seen, its corresponding features might still be implicitly represented within the feature sets of other ophthalmic diseases.
Specifically, we selected four disease categories that were entirely absent from our training set for testing: Retinitis Pigmentosa (RP), Hypertensive Retinopathy (HT), Macular Hole (MH), and Pathological Myopia (MYA). These categories were sourced from nine datasets that were completely held out during the training phase. As detailed under "Unseen Categories" in TABLE ~\ref{tab:dataset}, this zero-shot test set comprises two subsets: JSIEC (114 images) and ODIR (360 images).
The results, presented in Fig. ~\ref{fig:unseen}, clearly demonstrate the superiority of our model. Compared to the standard pre-trained CLIP model, our model achieves significantly higher accuracy in identifying these unseen categories. We also benchmarked against Vision Transformer (ViT) models that underwent Supervised Fine-Tuning (SFT). For a comprehensive comparison, we trained individual ViT models on each source dataset and then averaged their performance on this zero-shot task. The analysis revealed that such SFT models perform exceptionally poorly on zero-shot problems, likely due to overfitting to their source domains. Furthermore, other baseline models exhibited performance tantamount to random guessing on these unseen categories, failing to demonstrate any meaningful generalization capability.
In summary, these findings provide compelling evidence that our model, empowered by the predilection matrix, successfully constructs a semantically rich feature space. This allows it to transcend the explicit labels of the training data and achieve robust recognition of novel diseases.
\subsubsection{Ablation Study} 

We have validated the efficacy of not utilizing the OCT image encoder during the fine-tuning phase. We compared three distinct approaches for handling unpaired OCT images in the fine-tuning stage. The first approach involved randomly selecting one image from each category to serve as a weakly paired image for the fundus images. The second approach entailed employing the OCT encoder to encode all categories of OCT images into the latent space, subsequently calculating an average for each category. The third approach involved directly removing the OCT encoder module and utilizing the positional importance vector P as the learned spatial location information of diseases within the OCT latent space. As evidenced in Table~\ref{tab:estimation}, the method of randomly selecting a non-paired image from the same category significantly underperforms compared to the other methods. While averaging the latent space demonstrates some efficacy, it yields a lower PRC, indicating a susceptibility to the influence of sample size. For diseases with fewer samples, the averaged latent space fails to accurately represent the spatial locations of various diseases within the OCT space. The disease importance vector, learned through our approach, exhibits a marked advantage over the other methods.

\begin{table}[h]
\centering
\caption{Ablation Study on Various Processing Methods for Unpaired OCT}
\label{tab:estimation}
\resizebox{.98\linewidth}{!}{
\begin{tblr}{
  cell{1}{1} = {r=2}{},
  cell{1}{2} = {c=2}{c},
  cell{1}{4} = {c=2}{c},
  cell{1}{6} = {c=2}{c},
  cell{1}{8} = {c=2}{c},
  cell{1}{10} = {c=2}{c},
  cell{6}{2} = {c=2}{c},
  cell{6}{4} = {c=2}{c},
  cell{6}{6} = {c=2}{c},
  cell{6}{8} = {c=2}{c},
  cell{6}{10} = {c=2}{c},
  vline{2,4,6,8,10} = {1-9}{},
  hline{1,3,6} = {-}{},
  hline{7,10} = {1-9}{}
}
Models          & IDRID &       & APTOS2019 &       & Messidor &       & PAPILA &       & GF    &     \\
                & ROC   & PRC   & ROC       & PRC   & ROC      & PRC   & ROC    & PRC   & ROC   & PRC \\
RandomSelection & 0.725 & 0.501 & 0.876     & 0.619 & 0.819    & 0.529 & 0.714  & 0.535 & 0.905 & 0.792 \\
AverageLatent   & 0.784 & 0.577 & 0.920     & 0.675 & 0.842    & 0.560 & 0.758  & 0.610 & 0.925 & 0.861\\
LatentP         & 0.865 & 0.633 & 0.960     & 0.738 & 0.890    & 0.621 & 0.818  & 0.729 & 0.958 & 0.890\\
                & JSIEC &       & Retina    &       & RFMID    &       & ODIR   &       &       &     \\
RandomSelection & 0.854 & 0.842 & 0.901     & 0.793 & 0.871    & 0.480 & 0.835  & 0.503 &       &     \\
AverageLatent   & 0.934 & 0.902 & 0.935     & 0.864 & 0.925    & 0.552 & 0.887  & 0.615 &       &     \\
LatentP         & 0.997 & 0.972 & 0.945     & 0.877 & 0.940    & 0.576 & 0.919  & 0.721 &       &     
\end{tblr}
}
\end{table}

\section{Conclusions}
We propose a novel unpaired multimodal framework that uses OCT-derived spatial priors and contrastive learning to improve fundus image-based disease recognition. By dynamically learning disease-specific predilection sites and integrating them with fundus images, our model addresses modality imbalance and the high cost of paired multimodal data. The framework shows superior performance and generalizability across diverse datasets, offering a robust solution for ophthalmic disease identification.

\bibliographystyle{IEEEtran}
\bibliography{IEEEabrv,bibliography}

\end{document}